\documentclass[graybox]{svmult}

\usepackage{mathptmx}       % selects Times Roman as basic font
\usepackage{helvet}         % selects Helvetica as sans-serif font
\usepackage{courier}        % selects Courier as typewriter font
\usepackage{type1cm}        % activate if the above 3 fonts are
\usepackage{color}
\usepackage{cite}
\usepackage{makeidx}         % allows index generation
\usepackage{graphicx}        % standard LaTeX graphics tool
\usepackage{subfigure}
\usepackage{multicol}        % used for the two-column index
\usepackage[bottom]{footmisc}% places footnotes at page bottom
\usepackage{amsmath}
\usepackage[numbers]{natbib}
\usepackage{graphicx}
\usepackage{amssymb}
\makeindex             % used for the subject index
\begin{document}
	
	%\title*{A Procedure of Informative Path Planning Considering Wind-Field with Multiple UAVs for Measurement Mapping}
	\title*{Informative Path Planning and Mapping with Multiple UAVs in Wind Fields}
	\titlerunning{ Informative Path Planning and Mapping}	
	\author{Doo-Hyun Cho, Jung-Su Ha, Sujin Lee, Sunghyun Moon, and Han-Lim Choi}
	% Use \authorrunning{Short Title} for an abbreviated version of
	% your contribution title if the original one is too long
	\institute{Doo-Hyun Cho \at Department of Aerospace Engineering, KAIST, 291 Daehak-ro, Yuseong, Daejeon 34141, Korea, \email{dhcho@lics.kaist.ac.kr}
		\and Jung-Su Ha, Sujin Lee, Sunghyun Moon, and Han-Lim Choi \\ \email{\{jsha, sjlee, shmoon\}@lics.kaist.ac.kr, hanlimc@kaist.ac.kr}}
	%
	% Use the package "url.sty" to avoid
	% problems with special characters
	% used in your e-mail or web address
	%
	
	\maketitle
	
	\abstract{
\iffalse
		Informative path planning (IPP) is a planning problem for a robot to select the best path to gain information subject to given constraints.
		A key feature of this problem is that selecting locations to visit, dividing locations for each robot, and gathering information along the entire paths are strongly coupled.
		This paper proposes an systematic procedure of IPP for the measurement mapping with multiple UAVs: (a) select best locations to observe, (b) calculate the cost and find the best paths for each UAV, and (c) estimate the measurement value within the given region using the Gaussian process (GP) regression framework.
		Illustrative simulation result is presented to demonstrate the validity and applicability of the proposed approach.	
\fi
		Informative path planning (IPP) is used to design paths for robotic sensor platforms to extract the best/maximum possible information about a quantity of interest while operating under a set of constraints, such as dynamic feasibility of vehicles. The key challenges of IPP are the strong coupling in multiple layers of decisions: the selection of locations to visit, the allocation of sensor platforms to those locations; and the processing of the gathered information along the paths. This paper presents an systematic procedure for IPP and environmental mapping using multiple UAV sensor platforms. It (a) selects the best locations to observe, (b) calculates the cost and finds the best paths for each UAV, and (c) estimates the measurement value within a given region using the Gaussian process (GP) regression framework. An illustrative example of RF intensity field mapping is presented to demonstrate the validity and applicability of the proposed approach.	
%	The procedure adopts the mutual information and the Gaussian process regression framework to predict the input-output
	}

	\section{Introduction}
	\label{sec:1}
	Unmanned autonomous vehicles (UAVs) have been used as mobile sensor platforms for data collection in a variety of fields, including intelligence, surveillance, and reconnaissance missions, where the vehicles' mobility enables a large-scale sensing.
	Nonetheless, it can be difficult with just a single UAV to cover a huge area in a short time, particularly when the environment is changing rapidly.
	Also, from an economical perspective, data collection using multiple small UAVs with inexpensive onboard sensors is relatively cheaper and safer than carrying out the mission with a single UAV.
	
%	UAVs come in diverse shapes and sizes, but typically they are divided into two different categories; fixed wing and rotary wing.
%	Both platform possess its own key characteristics, and fixed wing types have simpler structure with more efficient aerodynamic performance which give longer flight durations and higher speeds.
%	These advantages make the fixed wing UAVs utilized in ISR missions in large area more frequently than rotary winged ones.
%	Although it is assumed to use fixed wing UAVs in this paper,

%	\textcolor{red}{
%	Several information measurement criteria have been used for informative path planning (IPP) problem formulation, such as fisher information, Kulback-Leibler divergence (relative entropy), uncertainty, and mutual information.
%	Levine\cite{levine2010information} used fisher information, and information-rich RRT (IRRT) algorithm was suggested to maximize the information acquisition which consideration of environmental and dynamic constraints.
%	Kullback-Leibler divergence is employed in Lu's work\cite{lu2014autonomous} in order to optimize the information value obtained by vision sensor while tracking multiple moving targets.
%	Binney et.al.\cite{binney2012branch} used the Gaussian process model to represent the uncertainty of interested region and used branch and bound method to obtain the best points to visit for informative path planning problem.
%	Using mutual information, Choi et.al.\cite{choi2010continuous} suggested a methodology of planning continuous paths for mobile sensors to reduce uncertainty of interested region in the future time domain.
%}
	
	Path planning for UAV missions is one of the key methods of ensuring efficient information acquisition, measurement, and mapping of a region of interest.
	Path planning problem have been studied for a long time, and most of that research has focused on minimizing the path length or the moving cost of the robot.
	However, in most cases, the highest priority of UAV path design should be to make the best use of the sensor and maximize the amount of information acuired in a given mission.
	Especially considering that the value or quantity of information is not equally distributed at every point in most cases, the planned path should prioritize areas that are information-rich.
	
	At the same time, one of the main constraints in path generation when operating a UAV system is the limited power source.
	In multiple studies, it has been shown that exploiting wind-energy is one of the most effective ways of decreasing UAV energy consumption ~\cite{ware2016analysis, langelaan2011wind, al2013wind}.
	Especially in mountainous regions, the wind blows strongly and can have a strong influence on the system dynamics of UAVs.
	This raises the question of whether a wind field can be exploited to operate the UAV system in a more efficient way.
	
	Several information measurement criteria have previously been used to formulate informative path planning (IPP) problems, such as the Fisher information~\cite{levine2010information}, the Kullback-Leibler divergence (relative entropy)~\cite{lu2014autonomous}, the Gaussian process uncertainty~\cite{binney2012branch}, and the mutual information~\cite{choi2010continuous}.
	Of these, we borrowed the meaning of information from ~\cite{shannon2001mathematical}, and the concept of `mutual information' with entropy is used in the following sections.
	In papers dealing with IPP problem information is generally defined as a signal obtained by missions, such as reconnaissance about a region or targets of interest.
	Also, there exist some studies (\cite{singh2009nonmyopic, lim2016adaptive}) which use an adaptive path planning in IPP problem since the non-adaptive setting of the problem is known as NP-hard \cite{singh2007efficient, guestrin2005near}, the non-linear solver was used in this study to avoid the local optimal solution.
	
	In this study, we focus on a mission which involves mapping the magnitude of measured data (e.g. an RF signal) in a specific region using a set of fully connected UAVs which are affected by wind.
	It is assumed that the mathematical model of the collected data is unknown, and therefore a probabilistic model, a Gaussian Process, is applied to calculate an estimate~\cite{rasmussen2006gaussian}.
	We develop a procedure for online mapping using path planning for multiple UAVs which maximizes the acquisition of information.
	Although there exist many studies involving the informative path planning problem, all of them were focusing on the partial side of the problem and there does not exist any studies which presented the whole procedure of the problem as mentioned above.
	
	Most of the studies about IPP have formulated target visitation to acquire information as a constraint~\cite{krause2008near} or a reward~\cite{binney2012branch}.
	Depending on the original intention, the obtained information can be used for classification of the target or for a decision making process.
	But that is beyond the scope of this paper, and here we only focus on the acquisition of the information.

%	The remainder of this paper is organized as follows.
%	In the section \ref{sec:prob_form}, a formulation of the problem handled in this paper is provided, and section \ref{sec:proc} shows the procedure of the problem in detail.
%	To verify the suggested procedure, simulation results are presented in section \ref{sec:num_ex}.
%	Finally we draw conclusions and discuss future work in section \ref{sec:conc}.
	
	% \vspace*{-0.1in}
	\section{Problem Formulation}
	\label{sec:prob_form}
	The problem to be handled in this paper is as follows: given a set of initial locations of UAVs in a bounded non-convex region with a wind field exerting an influence on the UAV dynamics, generate a path set which traverses every sensing points with minimum cost, and finally have all of the UAVs return to their initial locations.
	After the path set is generated, the UAV creates a map of the information to be measured along the path.
	Since the scope of the problem is wide and can be considered obscure, we subdivide the problem into 4 subproblems:
	
	\begin{enumerate}
		\item Given the number of tasks, optimize the set of task locations (representative spots to visit) for the maximum acquisition of information within the region of interest. (Subsection \ref{subsec:task_loc_opt}, eq. \eqref{eq:obj_tasklocation}.)
%		 For the simplicity, a region is defined in a 2-dimensional space.
		\item Calculate the paths and moving costs of a UAV between every pair of task locations obtained from subproblem $ \# $1. Because of the wind field dynamics, the paths and costs for both directions should be calculated separately. (Subsection \ref{subsec:calc_moving_cost}, eq. \eqref{eq:moving_cost}.)
		\item Using the output of the subproblem $\#$2, determine the minimum cost route which traverses all of the task locations and finally returns to the starting point. Since the situation is a multiple-UAV case, this subproblem can be modeled as a Multiple-Depot Multiple Traveling Salesmen Problem (MDMTSP). (Subsection \ref{subsec:mmmdmtsp}, eqs. \eqref{eq:costsum}, \eqref{eq:c_max})
		\item Each UAV obtains an information along the generated path. During travel, a map of the information to be measured is created for the region of interest. (Subsection \ref{subsec:GPR}, \eqref{eq:GPR1})
	\end{enumerate}
	
	Below are the assumptions that outline the constraints used in the problem.
	
	\begin{itemize}
		\item The energy consumption of each UAV is affected by the wind field dynamics.
		\item The UAVs are fixed wing and homogeneous; all of them share the same dynamics.
		\item The wind field dynamics and amount of information are stationary within a given region.
		\item The wind field dynamics is known in advance.
		\item Sensing is carried out by an onboard omni-directional sensor (e.g., radar, sonar) mounted on the UAVs, discretely along the paths (discrete time measurement).
		\item After sensing is finished, the UAVs return to their initial location.
	\end{itemize}
	
	% \vspace*{-0.1in}
	\section{Procedure}
	\label{sec:proc}
	
	% \vspace*{-0.1in}
	\subsection{Task Location Optimization based on Entropy and Mutual Information with Gaussian Random Variables}
	\label{subsec:task_loc_opt}
	
	% \vspace*{-0.1in}
	\subsubsection{Entropy and Mutual Information}
	Entropy, which measures the amount of uncertainty, in a random variable (R.V.) $ X $ with probability mass function $ p_{X} (x)$ is $ H(X) = -\mathbb{E} \left[\log(p(x))\right] $ (\cite{cover2012elements}).
	It can be also said that $ H(X) $ is approximately equal to how much information the R.V. $ X $ has on average.
	For multiple R.V.s, the joint entropy can be derived from the above definition and is $ H(X_1,X_2,\cdots,X_n) = \sum_{i=1}^{n}H(X_i|X_{i-1},\cdots,X_1). $
	Specifically, if a set of R.V.s follows the multivariate Gaussian distribution with mean $ \mu $ and covariance matrix $ K $, $ \textbf{X} \sim \mathcal{N}(\mu, K) $, then $ H(\textbf{X}) = \frac{1}{2} \log(2\pi e)^n |K| $
	where $ |K| $ denotes the determinant of $ K $ and $ n $ for the dimension of $ \textbf{X} $.
	
	The mutual information is the relative entropy (of KL-divergence) between the joint distribution and the product distribution between R.V. $ X_1 $ and R.V. $ X_2 $,
	\[
		I(X_1;X_2) = \mathbb{E}\left[\log \dfrac{p(X_1,X_2)}{p(X_1)p(X_2)}\right]
		= H(X_1) - H(X_2).
	\]
	It is also said that $ I(X_1;X_2) $ is a measure of the amount of information that one R.V. $ X_1 $ contains about another R.V. $ X_2 $, and can be interpreted as the reduction of entropy $ X_1 $ by conditioning on $ X_2 $.
	The mutual information for two sets of R.V.s which follow the multivariate Gaussian distribution, $ \textbf{X}_1 \sim \mathcal{N} (\mu_1, K_1)$, $ \textbf{X}_2 \sim \mathcal{N} (\mu_2, K_2)$, and $ (\textbf{X}_1,\textbf{X}_2) \sim \mathcal{N} (\mu, K)$, can be written as
	\begin{equation}\label{eq:MI}
		I(\textbf{X}_1;\textbf{X}_2) = \dfrac{1}{2}\log \dfrac{\det(K_1)\det(K_2)}{\det(K)}.
	\end{equation}

	% \vspace*{-0.1in}
	\subsubsection{Task Location Optimization}
	\label{subsubsec:task_loc_opt}
	To solve the first subproblem mentioned in Section \ref{sec:prob_form}, assume the interesting region $ A \subset \mathbb{R}^2 $ and the number of task locations $ n $ are given.
	The optimal set of locations $ P_T^* \subset A $ for a task set $ T = \{1,2,\cdots,n\} $ which maximizes the mutual information, $ I(\textbf{X}_T (P_T);\textbf{X}_o (P_o)) $, is (\cite{choi2014information, choi2010continuous})
	\newcommand{\argmax}{\operatornamewithlimits{argmax}}
	\begin{equation}\label{eq:obj_tasklocation}
		P_T^* = \argmax_{P_T \in A} I(\textbf{X}_T (P_T);\textbf{X}_o (P_o)).
	\end{equation}
	Here, $ \textbf{X}_T(P_T) $, the function of $ P_T $, is termed the \textit{verification variable} which represents the variables (or data set) obtained from the set of task locations $ T $, where $ \textbf{p}_t = (x_t, y_t) \in P_T $ is the location of $ t^{\textrm{th}} $ task, and $ t\in T $. $ \textbf{X}_o $ is a set of variables from the points called \textit{test points}.
	To calculate the entropy and the mutual information for the whole region of interest, locations for the \textit{test points} are equally spaced inside the region $ A $, and denoted as $ P_o $.
	The verification variable $ \textbf{X}_T $ is obtained from the set of locations $ P_T $, and $ \textbf{X}_o $ is obtained from $ P_o $.
	In this study, eqs. \eqref{eq:MI} and \eqref{eq:obj_tasklocation} were taken into the nonlinear programming solver\footnote{MATLAB \textit{fminsearch} function was used in this study.} to obtain the set of optimal points $ P_T^* $.
	The number of task locations $ n $ depends on the effective range and the noise level of the sensor.
	
	% \vspace*{-0.1in}
	\subsubsection{Meaning of Assigning Task Locations}
	\begin{figure}[t]
		\centering
		\begin{subfigure}
			\centering
			\includegraphics*[width=.8\textwidth, height=4cm, trim = 50 15 50 0]{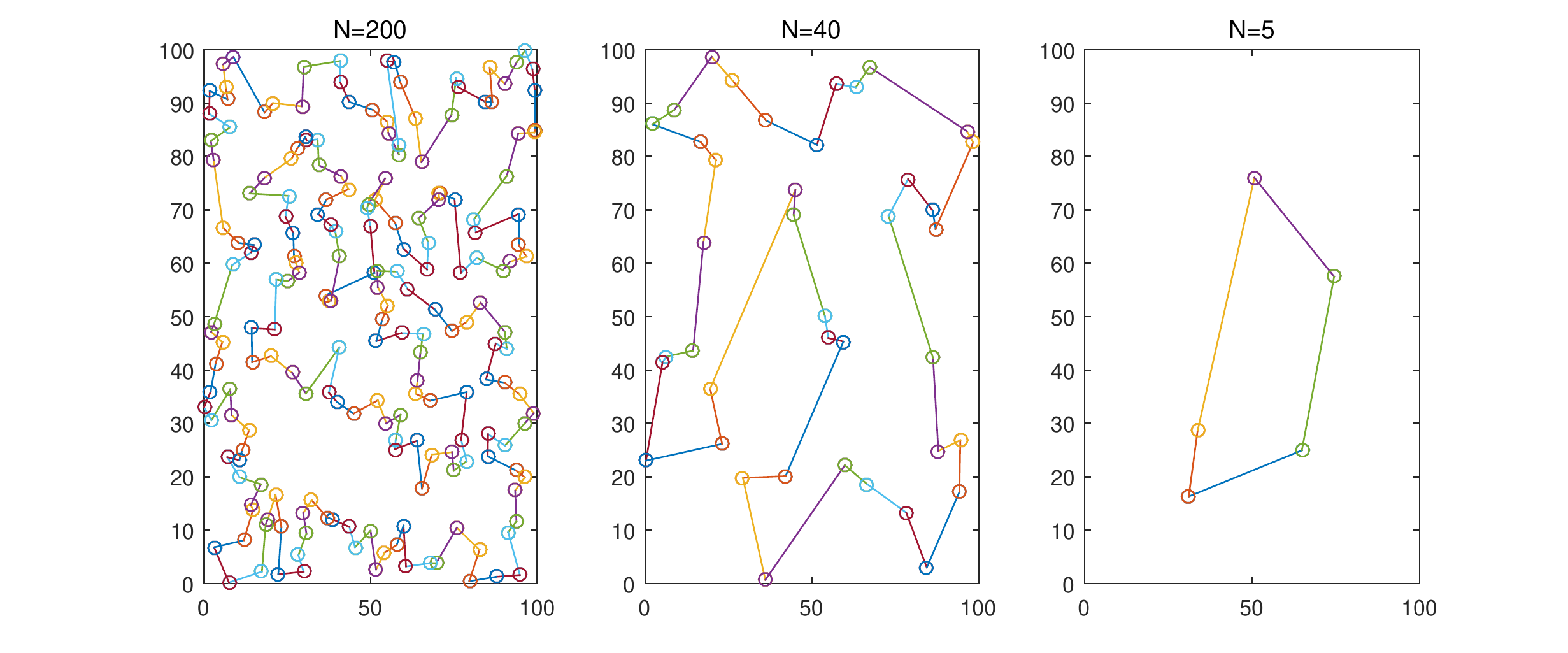}
			\caption{Randomly generated task locations with Hamiltonian path. Left : $ \# $Tasks : 200, Mid : $ \# $Tasks : 40, Right : $ \# $Tasks : 5. Left plot shows the path is overfitted, and right is underfitted.}
			\label{fig:path_vs_numtask}
		\end{subfigure}
		\begin{subfigure}
			\centering
			\includegraphics*[width=.7\textwidth, height=3.6cm, trim = 50 0 50 20]{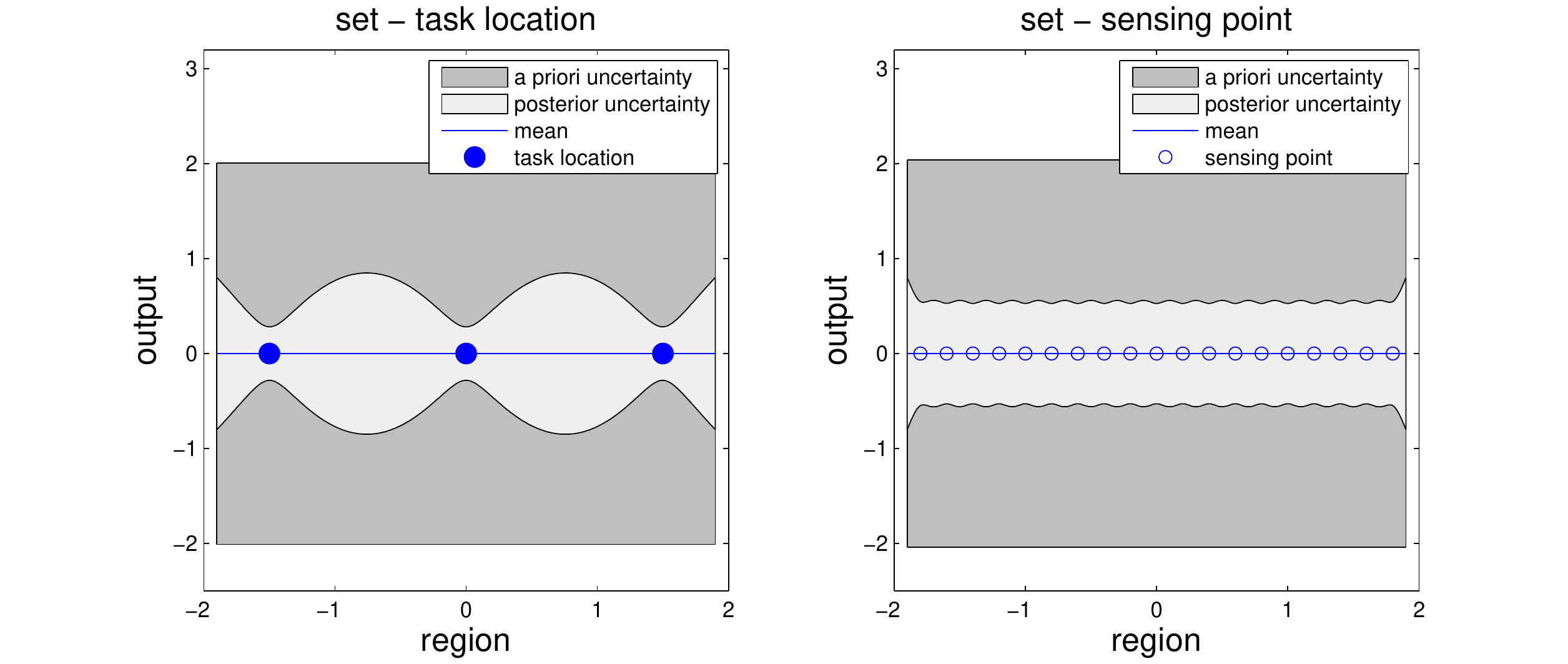}
			\caption{Uncertainty difference between before and after sensing. Left plot shows the concept of a role of task locations, and right plot shows the uncertainty level decreases with sensing.}
			\label{fig:meaning_of_task_location}
			\vspace*{-.25in}
		\end{subfigure}
	\end{figure}
	
	The set of optimal task locations, $ P_T^* $, is considered to be the set of representative spots, where the maximum amount of information can be obtained when the UAV tours around those locations.
	Suppose a set of task locations are given, and a UAV moves along the path which traverses all of the given locations.	
	Since it is assumed that the UAV is gathering information discretely during a tour even when it is not near one of the task locations, it is important to set the number of task locations properly for the path to be set well.
	In Fig. \ref{fig:path_vs_numtask}, the tasks with the Hamiltonian path in the region $ A \in [0, 100]\times[0, 100] $ are drawn for each case - the number of tasks $ N=$ 200, 40, and 5.
	Here the figure shows three cases of the generated path, \textit{overfitted, normal,} and \textit{underfitted}.
	If the number of tasks is set too high, then a region of overlapped sensing exists which is superfluous. Costs of moving increase unnecessarily, and we define this situation as \textit{overfitted}.
	Or if the number of tasks is too small to cover all of the region, the information obtained will be insufficient to estimate the data of interest after finishing the mission.
	We define this situation as \textit{underfitted}.
	
	The noise level and the length scale of a system model are important factors when setting the proper number of tasks for the mission.
	Described below is the relationship between the number of task locations and the number of sensing points.
	Assume that if the effective range of the sensor is very wide compared to the region of interest, then only few sensing points are enough to cover the whole region and there would be no problem to consider the task locations as the sensing points.
	However, in most of the cases, the inverse situation is given, and a lot of sensing points are needed to obtain the information of the whole region of interest.
	In this situation, if the number of task locations is given as the number of sensing points as mentioned above, and each UAV has to visit every task locations, then the \textit{overfitting} situation happens naturally.
	To avoid the \textit{overfitting} situation, the number of task locations should be less than the number of sensing points, and the proper number depends on the measuring frequency and the effective range of the sensor.
	
	To reduce the gap between the amount of mutual information obtained only on the task locations and the sensing points along the generated path for each UAV, it is necessary to assume that the noise level is less and the length scale is longer than the effective sensor range when optimizing the locations of tasks.
	Let $ MI_1 $ be the expected amount of mutual information gathered only from optimal task locations with low noise level and wide sensor range, and $ MI_2 $ be the expected amount of a mutual information gathered along the path by sensors mounted on a UAV.
	Using the above procedure mentioned in subsection \ref{subsubsec:task_loc_opt} with proper leveling, the difference between the amount of $ MI_1 $ and $ MI_2 $ converges to 0.
	The amount of mutual information can be interpreted as the difference between \textit{a priori uncertainty} and \textit{posterior uncertainty}, and this is indicated by the gray field in Fig. \ref{fig:meaning_of_task_location}.
	In the plots, the blue dots represent the verification variables with equally distributed locations in the 1-dimensional region, and the output was set to be 0.
	The output can be regarded as the measurement value of the sensor.
	We assume that the variables follow a Gaussian distribution with a linear mean function and a non-isotropic squared exponential covariance function. Details are shown in subsection \ref{subsec:GPR}.

	\subsection{Calculation of Moving Cost with FMT* algorithm}\label{subsec:calc_moving_cost}
	In order to compute the distances between every pair of task locations while considering the wind field, we adopted the Fast Marching Tree star (FMT*) algorithm which was recently proposed by Janson et al. \cite{janson2015fast}.
	The FMT* algorithm is a batch processing sampling-based path planning algorithm and it performs a direct dynamic programming process and lazy collision checking which dramatically accelerates the speed of the algorithm.
	
\begin{figure}[t]
		\centering
		\subfigure[]{
			\includegraphics*[width=.31\textwidth]{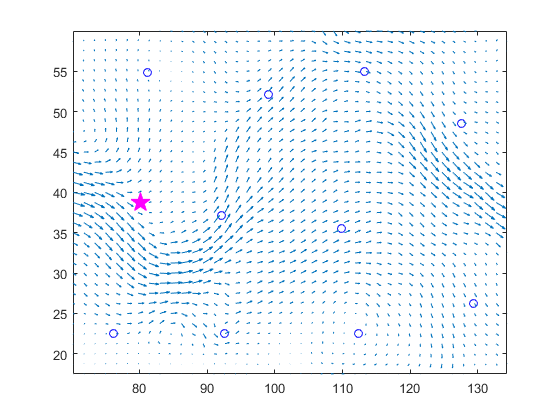}}
		\subfigure[]{
			\includegraphics*[width=.31\textwidth]{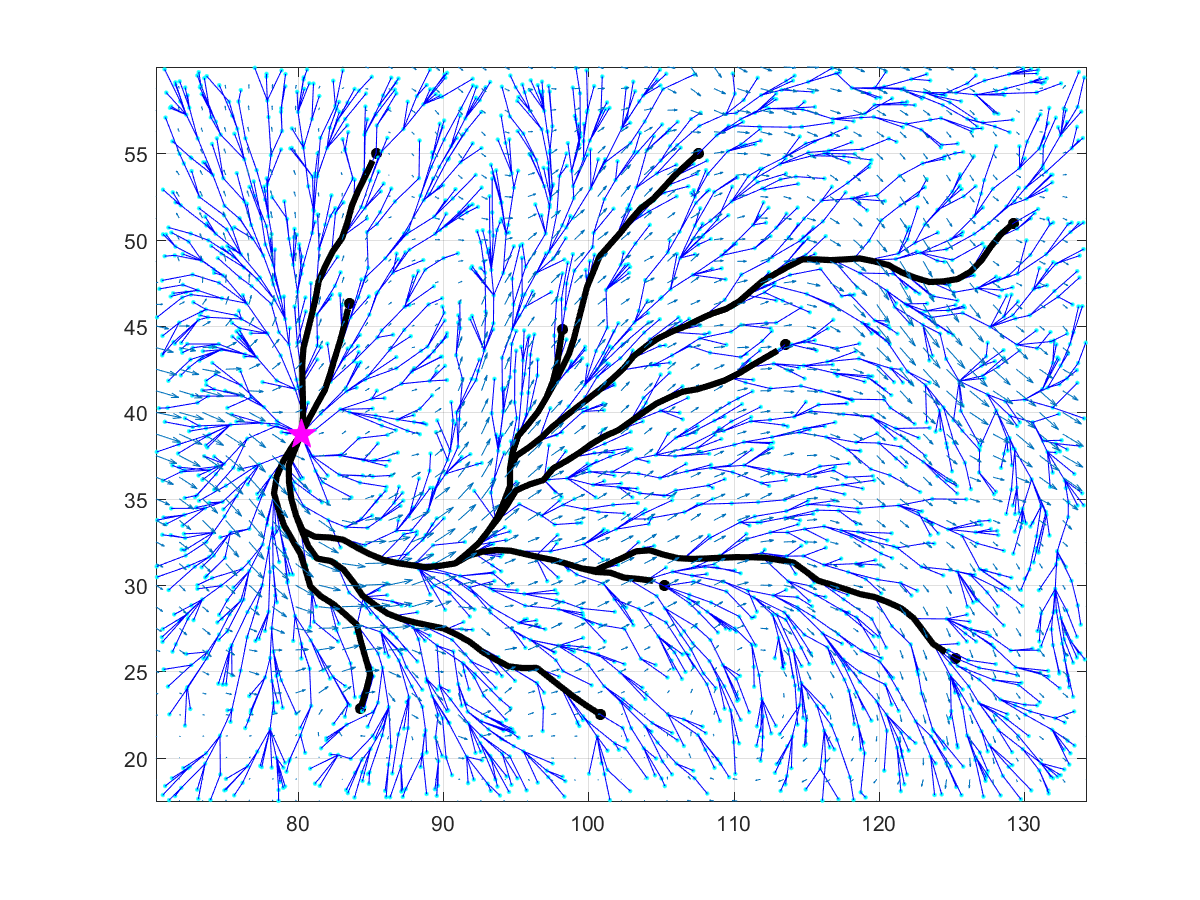}}
		\subfigure[]{
			\includegraphics*[width=.31\textwidth]{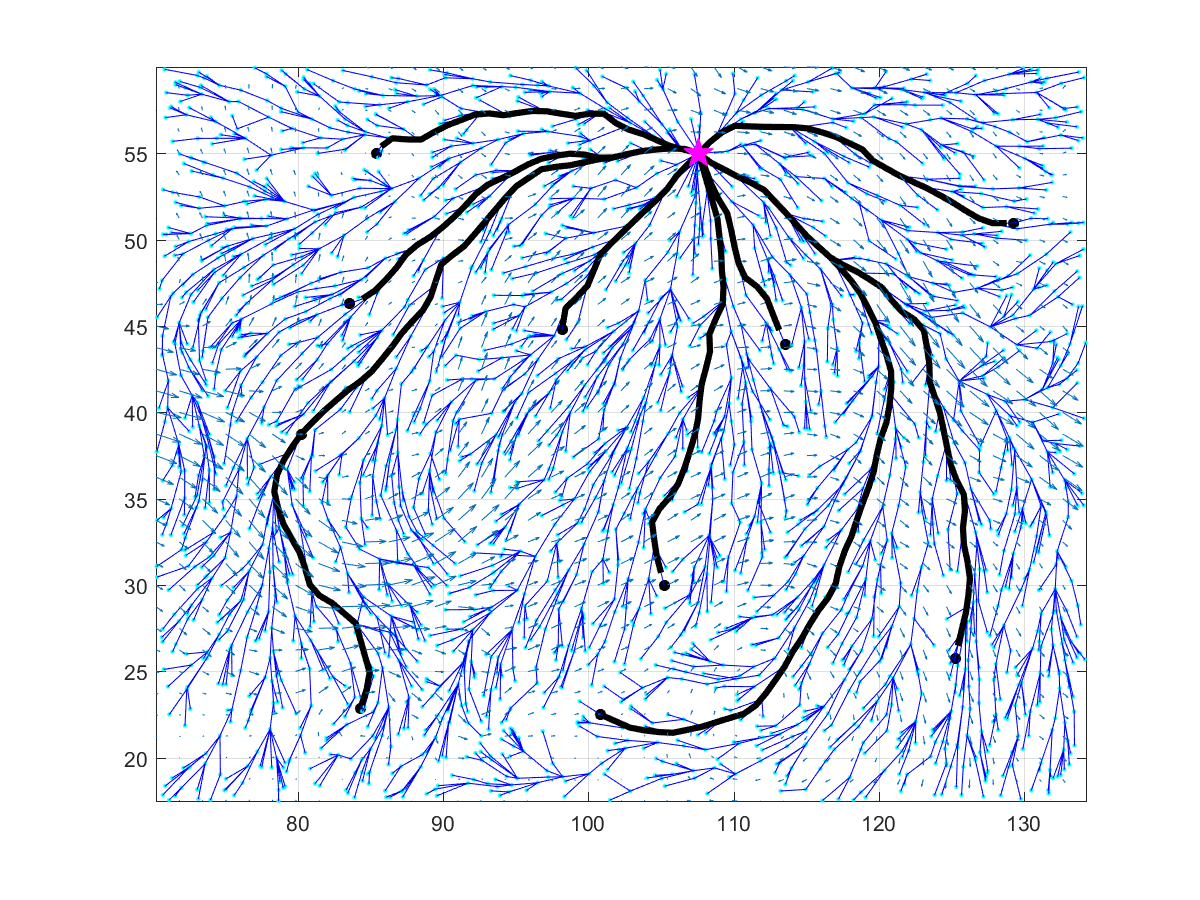}}
%		\subfigure[]{
%			\includegraphics*[width=.45\textwidth]{fig/3.png}}
		\caption{(a) Wind field and task locations. (b)-(c) Some resulting paths. Magenta stars and black lines represent starting locations and resulting paths, respectively.}
		\label{fig:result1}
	\end{figure}

	The algorithm was modified to be suitable for the problem stated in this study.
	First, while assuming that the UAVs follow the planned path, $\sigma:[0,\tau]\rightarrow\mathbb{R}^2$ (where $\tau$ denotes the duration of the path), with constant speed, $v_0$, the effect of the wind field, $\mathbf{w}\in\mathbb{R}^2$, was incorporated in the cost of the path planning problem.
	For the path direction, suppose that the UAV has a simple 2nd order dynamics with external force and drag as:
	$$
	\dot{v}(t) = \frac{F(t)}{m} - k(v(t)-w_\sigma),
	$$
	where $F,~m$ and $k$ denote the force, mass and drag coefficients of the UAV, respectively;
	$w_\sigma = \mathbf{w}(\sigma(t))\cdot\frac{\dot{\sigma}(t)}{||\dot{\sigma}(t)||}$ represents wind speed along the path direction.
	In order to maintain the constant speed, $v_0$,  the required force is given as $F_0(t) = mk(v_0-w_\sigma)$.
	Then, we obtained the total energy consumption along the trajectory as:
	\begin{align}
	E(\sigma) &= \int_0^\tau F_0(t) d||\sigma(t)|| \nonumber\\
					&= \int_0^\tau \left(mkv_0 - mkw_\sigma\right) d||\sigma(t)|| \nonumber\\
					&= mkv_0^2\tau - mk\int_0^\tau\mathbf{w}(\sigma(t))\cdot d\sigma(t),
	\end{align}
	where we assume $v_0>w_\sigma$ so that $F_0(t)>0$ for simplicity.
	The cost function of the path planning problem is obtained by simplifying the energy equation $C(\sigma) = E(\sigma)/mkv_0^2$:
	\begin{equation}\label{eq:moving_cost}
		C(\sigma) = \tau - \frac{1}{v_0^2}\int_0^\tau \mathbf{w}(\sigma(t))\cdot d\sigma(t).
	\end{equation}
	Note that there is a trade-off between path length and path direction: the cost penalizes the longer trajectory for the high desired speed $v_0$, while the path direction is encouraged to be aligned with the wind direction for lower $v_0$.
	Second, rather than solving the planning problem for every pair of task locations individually, we modified the algorithm into a multi-query version: with a fixed starting location, one planning problem finds all the paths to the other locations.
	As a result, if there are $n$ task locations, only $n$ (rather than $n^2$) planning problems need to be solved.
	Finally, we made all of the $n$ planning problems share the edge information.
	This was done because, in general, cost evaluation and the collision checking of edges are computational bottlenecks for the planning algorithm.
	Sharing edge information through the problems significantly improves the scalability of the algorithm.
	Fig. \ref{fig:result1} shows some of the resulting paths in an environment with an arbitrary wind field.
	It can be seen that the overall directions of the resulting paths are aligned with those of the wind field.
	
	\subsection{Mathematical Formulation of the Min-Max Multiple Depots Multiple Traveling Salesmen Problem (MMMDMTSP)}
	\label{subsec:mmmdmtsp}
	
	The MMMDMTSP~\cite{kivelevitch2012binary} is defined with a set of task locations $ T=\{1,2,\cdots,n\} $ and a set of depot locations (the initial location of the UAVs) $ D=\{n+1,n+2,\cdots, n+m\} $.
	The cardinality of $ T $ and $ D $ is denoted as $ |T|=n $ and $ |D|=m $, and it is assumed that $ n,m \geq 1 $.
	In this study, we let each UAV have a separate location, so if there are multiple vehicles in a depot, then the location of the depot will be repeated with how ever many vehicles exist in there. Therefore, $ m $ also represents the number of UAVs.
	
	Let $ G=(V,E) $ be a directed graph where $ V=T\cup D =\{1,2,\cdots,n,n+1,\cdots,n+m\}$ is the union of sets of the task locations and depot locations, and $ E=\{(i,j) | \forall (i,j) \in (V\times V) \backslash (D\times D)\} $ is the edge set (note that $ E $ doesn't include any edge between depots) with the cost denoted by $ c_{ij} $.
	The reason that $ G $ is not an undirected graph is that the cost from $ i $ to $ j $, $ c_{ij} $ and the cost from $ j $ to $ i $, $ c_{ji} $ are different because of the wind field dynamics model.
	
	For each edge, $ \theta_{ijk}\in \{0,1\} $, is defined to be equal to one if UAV $ k $ takes the edge $ e=(i,j) $ as a part of its route and zero otherwise, where $ i,j\in V, i \neq j $, and $ k\in D $. 
	To prevent a UAV taking a loop for a certain task location, $ \theta_{iik}=0 $ should be added as a constraint.
	
	From the above definitions, the cost sum of the $ k $th UAV route can be formulated as:
	\begin{equation}\label{eq:costsum}
		C_k = \sum_{j=1}^{n}\theta_{kjk}c_{kj} + \sum_{i=1}^{n}\sum_{j=1}^{n}\theta_{ijk}c_{ij} + \sum_{i=1}^{n}\theta_{ikk}c_{ik}.
	\end{equation}
	The first term represents the cost from the $ k $th UAV's depot to the first task location. The second term is a sum of the costs along the route for set of task locations, and the final term is the cost from the last task location to the UAVs' depot.
	
	To minimize the longest tour for every UAV, $ C_{max} = \max_{k\in D}C_k $ is defined as a continuous decision variable, and the following is the formulation of the MMMDMTSP:
	\begin{equation}\label{eq:c_max}
		\textrm{Minimize} \quad C_{max}
	\end{equation}
	such that
	
%	\small
	\begin{align}
		\sum_{k=1}^{m}\theta_{kik} + \sum_{j=1}^{n}\sum_{k=1}^{m}\theta_{jik} = 1 & & \forall i\in N \label{eq:const1}\\
		\theta_{kik} + \sum_{j=1}^{n}\theta_{jik} - \left( \theta_{ikk} + \sum_{j=1}^{n}\theta_{ijk}\right) = 0 && \forall i,j \in T, \forall k \in D \label{eq:const2}\\
		\sum_{i=1}^{n}\theta_{kik}\leq 1 && \forall k \in D \label{eq:const3}\\
		\sum_{i=1}^{n}\theta_{kik} - \sum_{i=1}^{n}\theta_{ikk} = 0 && \forall k \in D \label{eq:const4}\\
		C_k \leq C_{max} && \forall k \in D \label{eq:const5}\\
		\sum_{i\in S}\sum_{j\in S}\theta_{ijk} \leq |S| - 1 &&
		\begin{matrix}
			i\neq j, \forall k\in D, \\
			|S| = 2,3,\cdots, n-1
		\end{matrix} \label{eq:const6}
	\end{align}
%	\normalsize
	
	The constraint \eqref{eq:const1} ensure that all the task locations are visited exactly once by the set of routes, and \eqref{eq:const2} guarantees that each task is not the final destination of the routes.
	Eqs. \eqref{eq:const3} and \eqref{eq:const4} are the constraints that respectively ensure that each UAV has at most one route, and if it does have more, then the UAV should return to its initial depot (this makes the problem a TSP(traveling salesman problem), not a VRP(vehicle routing problem)).
	Eq. \eqref{eq:const5} balances the cost of each route for every UAV. 
	Finally, \eqref{eq:const6} is a constraint for subtour elimination.
	Here, a subtour is a route which only has vertices $ i \in T $ and is not from the set of depots, $ D $~\cite{kivelevitch2012binary}.
	
	Since it is intractable to set up the subtour elimination constraint equation for every cases, the number of this constraint equation is a determining factor for solving this MILP formulation.
	The GA (Genetic algorithm) was adopted in this work, and we followed the details shown in \cite{ombuki2009using}.
	
	% \vspace*{-0.1in}
	\subsection{Gaussian Process Regression}
	\label{subsec:GPR}
	
	The Gaussian Process Regression \cite{rasmussen2006gaussian} (GPR) is one of the supervised modeling scheme that approximates the interesting target points (output points) using the function of training points (input points).
	This scheme regards the relationship between input and output points as one of the examples of the Gaussian process, and thus the output points can be approximated as a kind of Bayesian inference which computes the posterior distribution of output points of interest conditioned on the experimentally obtained input points.
	In particular, it is assumed that all of the relevant probability distributions of input and output points follow a joint Gaussian distribution.
	With input points, the GPR procedure consists of defining a mean function and a covariance function and learning their hyperparameters, to maximize the probability of the GPR generating interesting output points.
	Here the hyperparameters determine the shape and characteristics of the GPR.
	
	The most used mean function and covariance in GPR are described below; for the mean function, a constant function is enough to estimate the model in many cases:
	\begin{equation}\label{eq:mean}
		\mathbb{E}[f(\textbf{x})] = m(\textbf{x}).
	\end{equation}
	For the covariance function, the non-isotropic squared exponential covariance function is used:
	\begin{equation}\label{eq:cov}
		\mathrm{cov}\left(f(\textbf{x}_p), f(\textbf{x}_q)\right)
		= k(\textbf{x}_p, \textbf{x}_q)
		= \sigma_f^2 \exp \left[ -\dfrac{1}{2} (\textbf{x}_p - \textbf{x}_q)^T \Sigma^{-2}_l (\textbf{x}_p - \textbf{x}_q) \right]
	\end{equation}
	where $ \sigma_f $ is the signal's standard deviation, and $\Sigma_l=\mathrm{diag}\left(\sigma_{l_1}, \cdots, \sigma_{l_n}\right)$, which represents the characteristic length-scale for each input dimension.
	These parameters,  $\sigma_f$ and $\Sigma_l$, are called hyperparameters.
	
	In the Gaussian Process, for an arbitrary input point set $\{\textbf{x}_1, \cdots, \textbf{x}_n\} \in \textbf{X}$, the output point set $\{f(\textbf{x}_1), \cdots, f(\textbf{x}_n)\}$ follows the following Gaussian distribution.
	\[
	\left[
	\begin{array}{c}
	f(\textbf{x}_1) \\
	\vdots \\
	f(\textbf{x}_n)
	\end{array}
	\right]
	\sim\mathcal{N}
	\left(
	\left[
	\begin{array}{c}
	m(\textbf{x}_1) \\
	\vdots \\
	m(\textbf{x}_n)
	\end{array}
	\right]
	,
	\left[
	\begin{array}{ccc}
	k(\textbf{x}_1, \textbf{x}_1) & \cdots & k(\textbf{x}_1, \textbf{x}_n)  \\
	\vdots & \ddots & \vdots \\
	k(\textbf{x}_n, \textbf{x}_1) & \cdots & k(\textbf{x}_n, \textbf{x}_n)
	\end{array}
	\right]
	\right)
	\]
	
	In the case of predicting with noisy observations, it is known that the predictive equations for GPR are
	\begin{equation}
		\textbf{f}_* | X, \textbf{y}, X_* \sim \mathcal{N} \left( \bar{\textbf{f}}_*, \textrm{cov}(\textbf{f}_*) \right)
		\label{eq:GPR1}
	\end{equation}
	where $ \bar{\textbf{f}}_* = K_{OI}\left[ K_{II}+\sigma_n^2 I \right]^{-1}\textbf{y} $ and $ \textrm{cov}(\textbf{f}_*) = K_{OO}-K_{OI}\left[ K_{II}+\sigma_n^2 I \right]^{-1}K_{IO} $.
	Here $\textbf{y}$ is the observation vector with additive Gaussian noise.
	The subscript I indicates the input data of size I, and O is the interesting point to be verified.
	$K_{II}$ denotes the $I \times I$ matrix of the covariances evaluated at all pairs of input and input points, and similarly for the other matrices $K_{OI}$ and $K_{OO}$.
	
	As mentioned above, the hyperparameters determine the characteristics of the GPR structure; so an optimization process to obtain the proper hyperparameters is needed for an accurate approximation.
	The most commonly used method of optimization is to choose the hyperparameters which maximize the log likelihood of the given input points, in other words:
	\begin{equation}
		(\sigma_f,\Sigma_l)=\argmax_{\sigma_f,\Sigma_l}\log p(\textbf{y}|X)
		\label{eq:GPR2}
	\end{equation}
	
	In subproblem $ \# $4, an information map of an interesting region $ A $ can be constructed with GPR, \eqref{eq:GPR1} and \eqref{eq:GPR2}.
	From \ref{subsubsec:task_loc_opt}, the training points are $ P_T $ with the values $ X_T(P_T) $, and the target points are the set of locations $ P_o $ to obtain $ X_o(P_o) $.
	
	% \vspace*{-0.1in}
	\section{Numerical Example}
	\label{sec:num_ex}
	
	This section shows the detailed results of the simulation for the problem given in Section \ref{sec:prob_form}, using the suggested procedure (Section \ref{sec:proc}).
	The information to be measured is assumed to be the intensity of the RF signal, and a signal intensity map is generated with the obtained information.
	The parameters and conditions used in the simulation are as follows.	
	
	% \vspace*{-0.1in}
	\subsection{Simulation Parameters and Conditions}
	\label{subsec:sim_param_cond}
	
	% \vspace*{-0.1in}
	\subsubsection{Wind-Field}
	
	There are several kinds of wind field models that can be used for generating wind field data.
	Each model depends on the degree of simplification of the \textit{Navier-Stokes} equation, and this point determines the spatial resolution of the output data.
	It is necessary to choose the wind field model with the appropriate spatial resolution when the size of the region is given.
	In this simulation the size was fixed as $ 20km \times 20km $.
	The wind field data was generated with software called $ WindSim $, which is based on the \textit{Computational Fluid Dynamics} model, with a spatial resolution of $ 20m $, and this resolution size fits the assumed size of the interesting region.
	If the temperature is assumed to be constant, the streamline of the wind differs with the following factors: variation in the altitude of the surface, and land uses (forest, farm, downtown, mountain, etc.).
	To obtain wind field data, a digital elevation model (DEM), land uses, and the boundary conditions of the wind field are needed as inputs of the program.
	In the simulation the region of interest was assumed to be a mountainous area.
	The boundary condition of the wind was set to be 10 $ m/s $ from the northeast.
	The 3D wind field data was obtained with these inputs and conditions, and the data at specific altitude, which is the 2D data along the xy plane, was chosen to be used in the simulation below.
	In Fig. \ref{fig:path_gen}, the generated wind field is shown as a vector plot (blue arrow).
	
	% \vspace*{-0.1in}
	\subsubsection{RF Signal Intensity Data}
	
	The basic RF propagation model can be expressed using the \textit{free space path loss }model, $ P_L(dB)=\log_{10}\dfrac{P_t}{P_r} = -10\log_{10}\dfrac{G \lambda^2}{(4\pi d)^2}$ where $ \lambda $ is the signal wavelength, $ d $ is the distance from the transmitter, and $ G $ is the gain value.
	But a change in intensity always happens because of geographical features, and this effect is called \textit{shadowing}.
	It is usually assumed that shadowing follows lognormal distribution, and an RF signal intensity map can be generated for a region with complex geographical features using the \textit{combined path loss and shadowing} model \cite{hufford1982guide}.
	The program called \textit{Radio mobile} was used to obatin an RF signal intensity map of an interesting region at the frequency band of $ 146MHz $.
	The type of transmitter antenna was assumed to be omni-directional, and the gains of the transmitter/receiver antenna were 6.0/2.0 $ dBi $ respectively.
	It was assumed that one transmitter was in the center of the region.
	A signal intensity map of the region is shown in Fig. \ref{fig:rf_map}.
	The intensity plot is drawn for an altitude of $ 3,000 m $.
	
	% \vspace*{-0.1in}
	\subsubsection{Information Acquisition with Onboard Sensor}
	
	It was assumed that a total of 3 UAVs were used to obtain information in this simulation, and that each of them moved along the generated path inside the given region.
	The initial locations of each UAV can be arbitrarily chosen, but it was assumed that the locations were given far away moderately from each other.
	The altitude of the UAVs was fixed at 3,000$ m $.
	The UAV dynamics followed the simple Dubins path model, where $ \dot{x}=v \cos(\theta) $, $ \dot{y}=v \sin(\theta) $, and $ \dot{\theta}=u $ where $ u $ is bounded.
	The speed and the minimum turning radius were $ v = 100 m/s$, $ r_{min} = 50m $ each.
	To generate an RF signal intensity map using GPR, Eqs. \eqref{eq:mean}, \eqref{eq:cov} were used as the mean and covariance function.
	We set the hyperparameters as follows: $ \sigma_f = 30 $dBm, $ \sigma_n = 1 $dBm, and $ l_x, l_y = 4 $km.
	Each sensor measured the data for every 10 seconds, and during the data acquisition the optimization of these hyperparameters were carried out with Eq. \eqref{eq:GPR2} to maximize the amount of mutual information.
	$ \sigma_n $ is a sensor dependent value which is not the subject of a hyperparameter optimization.
	
	% \vspace*{-0.1in}
	\subsection{Simulation Result}
	\begin{figure}[t!]
		\centering
%		\begin{minipage}{.55\textwidth}
%			\centering
%			\includegraphics[width=\textwidth, height = 4cm, trim = 60 30 0 30]{fig/windfield_mountain.pdf}
%			\caption{Wind Field for UAV dynamics constraint.}
%			\label{fig:wind_field}
%		\end{minipage}
		\begin{minipage}{.4\textwidth}
			\centering
			\includegraphics[width=\textwidth, height = 3.7cm, trim = 30 0 0 0]{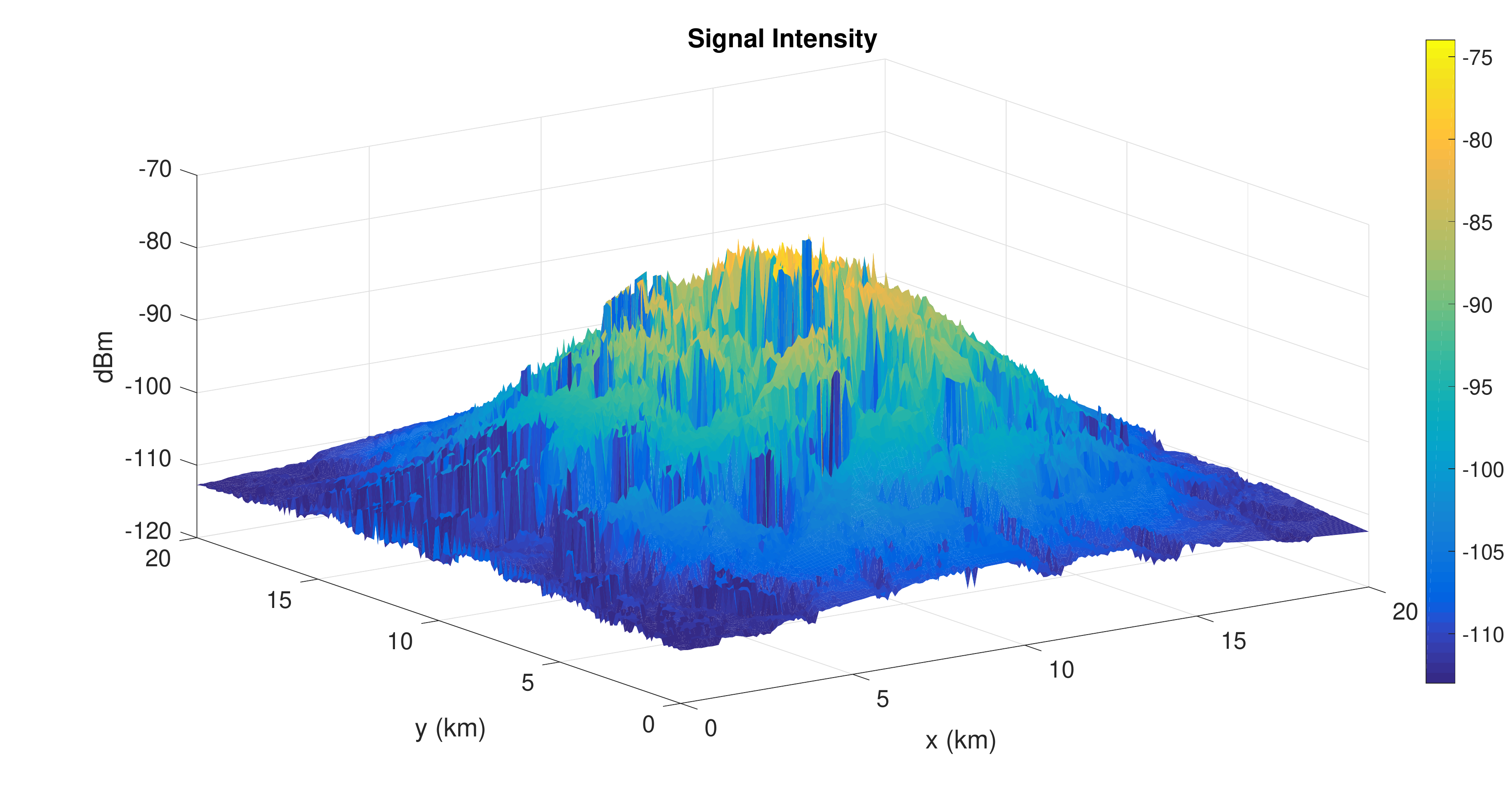}
			\caption{RF Signal Intensity with altitude 3$ km $. Unit : dBm.}
			\label{fig:rf_map}
			\vspace*{-.05in}
		\end{minipage}
		\hspace{.035\textwidth}
		\begin{minipage}{.35\textwidth}
			\centering
			\includegraphics[width=\textwidth, trim = 60 0 0 0]{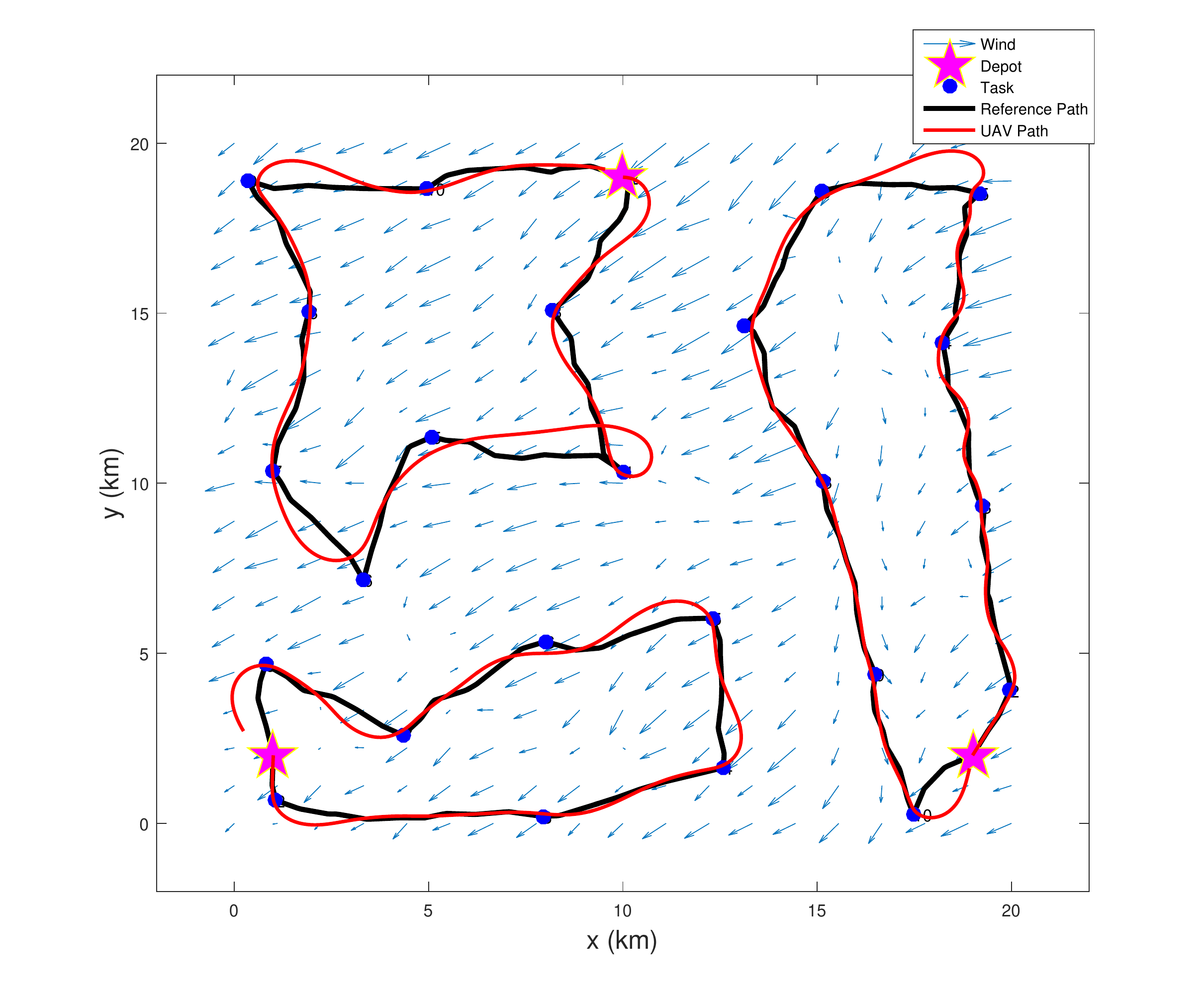}
			\caption{Locations of depots (red stars) and tasks (blue dots), reference paths (black line), and UAV paths (red line). }
			\label{fig:path_gen}
		\end{minipage}
		\begin{minipage}{\textwidth}
			\centering
			\subfigure[]{
				\includegraphics*[width=.4\textwidth, trim = 60 0 0 0]{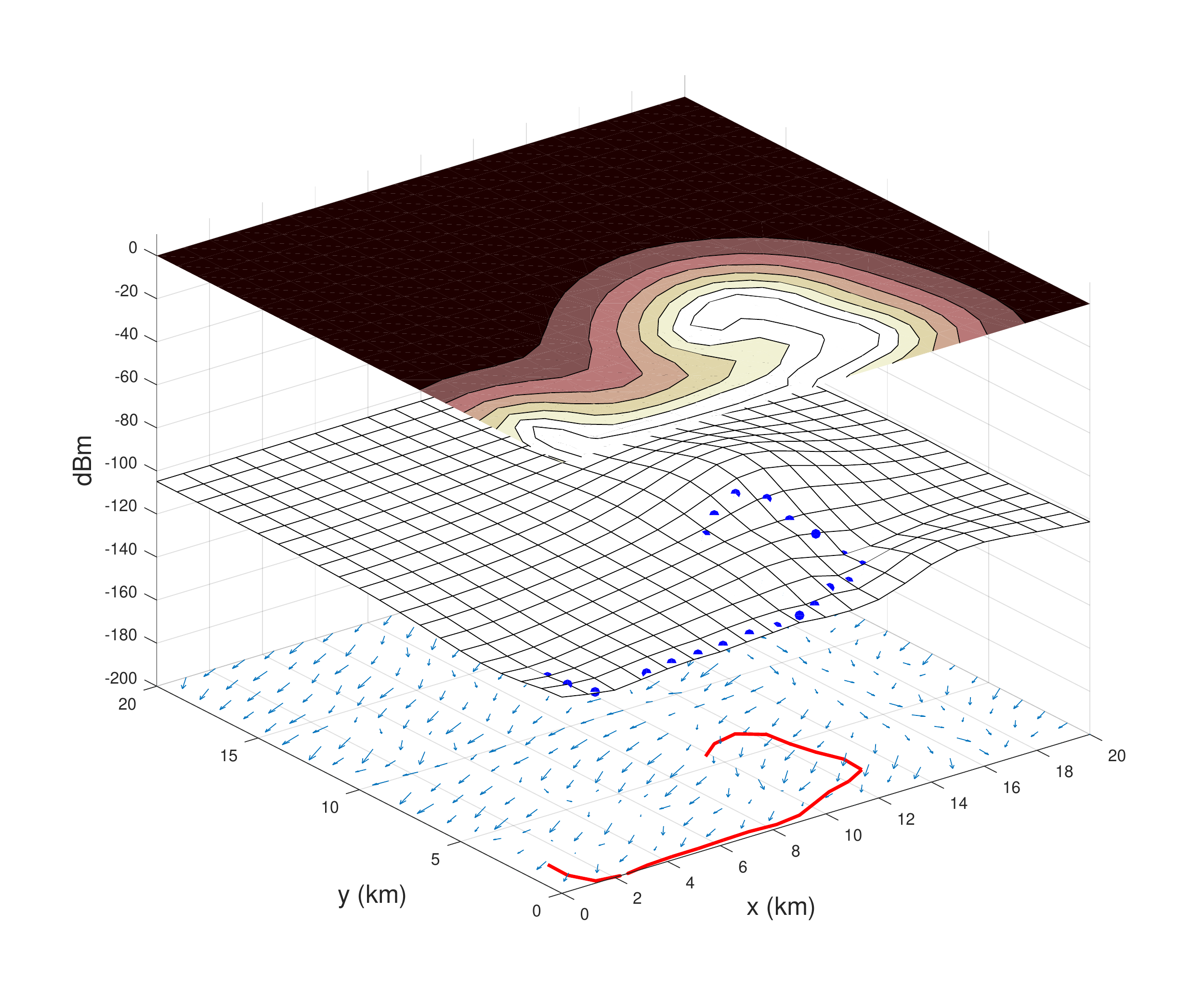}
				\label{fig:result_init}}
			\subfigure[]{
				\includegraphics*[width=.4\textwidth, trim = 60 0 0 0]{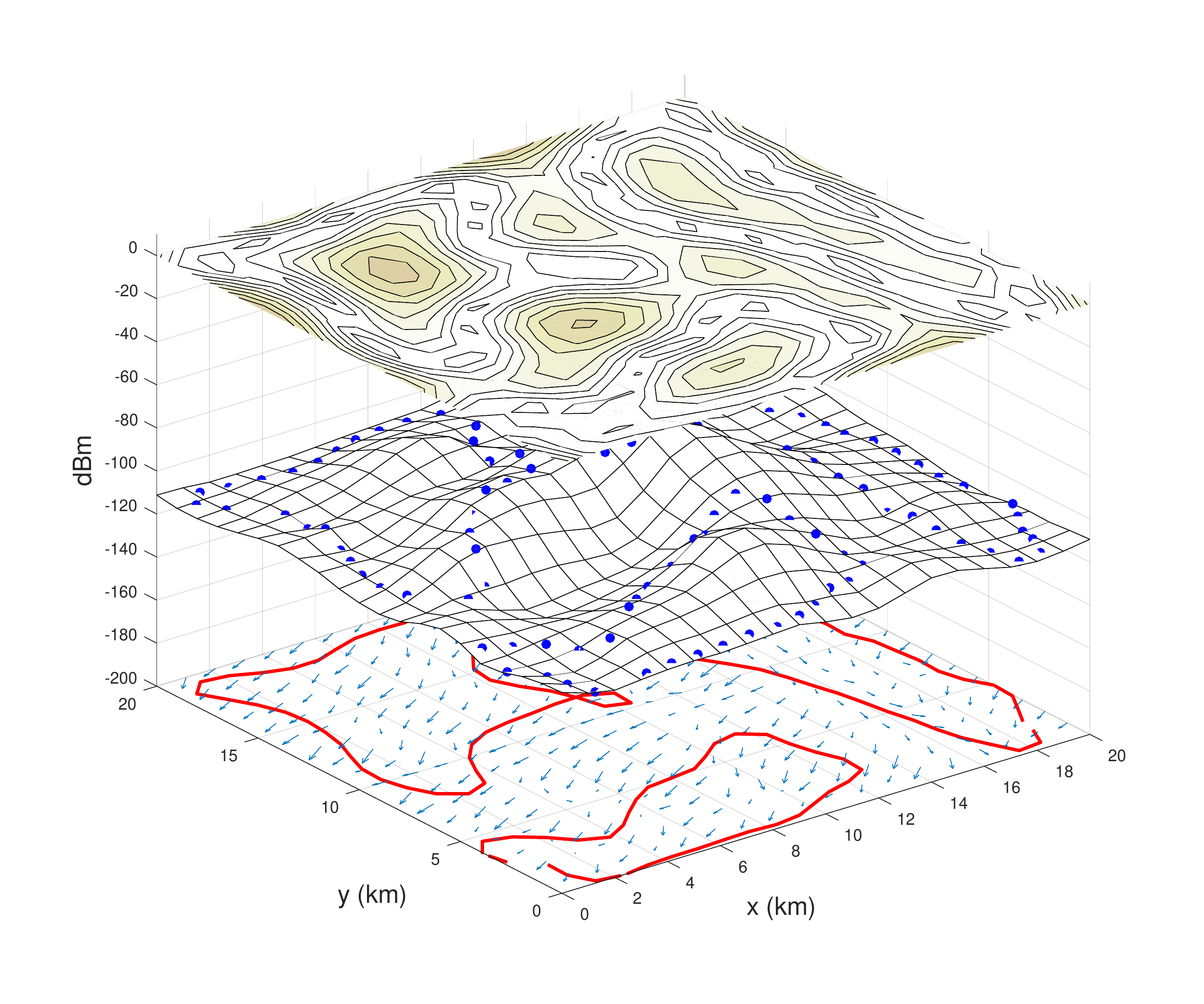}
				\label{fig:result}}
			\caption{Simulation Results. (a) Initial stage - High uncertainty level, low information gained. (b) Final stage - Low uncertainty level, high information gained. Top : The uncertainty level of the region of interest. Middle : The estimated RF map. Bottom : The portion of the paths executed by the UAVs}
		\end{minipage}
		\vspace*{-.2in}
	\end{figure}
%	\begin{figure}
%		\centering
%		
%	\end{figure}
	
	The results are shown in Figs. \ref{fig:path_gen}, \ref{fig:result_init}, and \ref{fig:result}.
	These were obtained with the parameters and conditions in \ref{subsec:sim_param_cond}.
	Fig. \ref{fig:path_gen} shows the results of the subproblems \#1 to \#3.
	Since it is assumed that the amount of information is equally distributed within the region, the task locations are spread apart from each other.
	The moving costs for every pair of tasks were obtained using the FMT* algorithm to construct a distance matrix.
	This matrix becomes an input of the MDMTSP problem, and the output of this problem (reference paths for each UAV) is drawn as a black line.
	The depot locations are marked as red stars; one is located in the top center, and the others are on the left and right bottom in the given region.
	Red lines are the paths generated by the UAV dynamics for the given reference paths.
	
	Figs. \ref{fig:result_init}, \ref{fig:result} show the mapping results for RF signal intensity and the level of uncertainty within the region.
	The bottom contour shows how far the UAVs have gone along the paths.
	The 3D plot in the middle is the estimated RF map using GPR.
	Each of the blue dots are the points where sensing is performed, and the mapping is based on this obtained data.
	The level of uncertainty is shown in the upper contour; a domain with bright color indicates that the uncertainty level is low (or the information has been obtained), otherwise if the color is dark, the uncertainty level is high.
	Fig. \ref{fig:result_init} shows the early stage of the mission, and the final result is shown in Fig. \ref{fig:result}.
	As the UAVs move and take measurements, the mapping becomes similar to the original data and the level of uncertainty gets lower.

	% \vspace*{-0.1in}
	\section{Conclusion}
	\label{sec:conc}
	
	An IPP procedure to measure and map a region of interest using multiple UAVs has been proposed, in the framework of mutual information and the Gaussian process regression.
	The validity of the procedure was demonstrated by simulation using a realistic wind field and RF signal intensity data, and the target region was assumed to be a mountainous area.
	
	In future work, a more realistic MDMTSP with Dubins path concept will be considered as well as various amounts and distribution of information inside the region.
	Also, the problem can be extended to time-varying situations, which makes the problem harder to analyze and solve.

	\begin{acknowledgement}
%		This research was supported by the \textcolor{red}{International collaborative work with Collorado University} funded by the Agency for Defense Development, Korea.
	This work was supported by Agency for Defense Development (contract \#UD140053JD).

	\end{acknowledgement}
	% \vspace*{-0.1in}
	% \vspace*{-0.1in}
	\small
	\bibliographystyle{spbasic}
	\bibliography{bibtex_database}
\end{document}